\documentclass[11pt]{article}

\usepackage{hyperref}
\usepackage{amsmath}
\usepackage{amssymb}
\usepackage[pdftex]{graphicx}
\usepackage[australian]{babel}

\usepackage{todonotes} 

\begin{document}



\title{Analysing Soccer Games \\ with Clustering and Conceptors}


\author{\protect\parbox{\textwidth}{\protect\centering Olivia Michael$^1$ \hspace{0.3cm} Oliver Obst$^1$  \\ Falk Schmidsberger$^2$ \hspace{0,3cm} Frieder Stolzenburg$^2$ }}
 
%
%
%
\date{\footnotesize
	(1) Centre for Research in Mathematics, Western Sydney University,\\ 
	Locked Bag 1797, Penrith NSW 2751, Australia\\
	EMail: {\tt 17974761@student.westernsydney.edu.au, o.obst@westernsydney.edu.au} \\[1em]
(2) Department of Automation and Computer Sciences, Harz University of Applied Sciences, Friedrichstr. 57-59, 38855 Wernigerode, Germany\\
        EMail: \tt{\{fschmidsberger,fstolzenburg\} at hs-harz.de}
	}

\maketitle

\begin{abstract}
We present a new approach for identifying situations and behaviours, which we call \emph{moves}, from soccer games in the 2D simulation league.
Being able to identify key situations and behaviours are useful capabilities for analysing soccer matches, anticipating opponent behaviours to aid selection of appropriate tactics, and also as a prerequisite for automatic learning of behaviours and policies.
To support a wide set of strategies, our goal is to identify situations from
data, in an unsupervised way without making use of pre-defined soccer specific
concepts such as ``pass'' or ``dribble''. The recurrent neural networks we use
in our approach act as a high-dimensional projection of the recent history of a
situation on the field. Similar situations, i.e., with similar histories, are
found by clustering of network states. The same networks are also used to learn
so-called conceptors, that are lower-dimensional manifolds that describe
trajectories through a high-dimensional state space that enable
situation-specific predictions from the same neural network.
With the proposed approach, we can segment games into sequences of situations that are learnt in an unsupervised way, and learn conceptors that are useful for the prediction of the near future of the respective situation. 

\end{abstract}

\section{Introduction}

Some of the recent achievements of AI systems~\cite{BS17,FBC+10,MKS+15,SHM+16} have in common that they use some form of machine learning as a key component, and that their performance can be immediately compared against the performance of a human player. Teams in the soccer simulation leagues play against each other, a comparison against humans is not a part of the competition.\footnote{To our knowledge, experimental games in the earlier years of RoboCup against a team of humans using joysticks to control players were easily won by computer programs. It should be noted, however, that the soccer simulation was not designed to be played by humans, and how human players interface with the simulation will obviously affect performance to a large degree.}
As such, it is impossible to say how well simulation league teams play in terms of potential human performance. It is, however, possible to observe continuing progress in performance of champion teams~\cite{GFG17,GR11}. 
In this paper, we propose a new method for automatically analysing soccer matches, a step we see as a prerequisite to learning behaviours and strategy for a team.

\subsection{Motivation}

A capability to model and describe possible situations and interactions in a game of autonomous robots has benefits for training or programming the team. With a meaningful vocabulary of situations, appropriate actions and strategies can be implemented, and together with a deeper analysis of past soccer matches, likely future situations can be better predicted. 

In the context of RoboCup soccer, several efforts have been made to support the
description of situations, interactions between players, or sequences of actions
and situations. In \cite{Mur04,MOS01b}, for example, UML statecharts are used to
describe plans of individual agents and the team, with the goal to manually
specify team behaviour. An information-theoretic approach for the analysis of
interactions has been developed in~\cite{CLW+17}; this approach is able to
identify dynamic relationships between players of a team and its opponents. In
contrast to the UML statecharts approach, the information-theoretic analysis of
interactions is model-free, and does not rely on predefined, soccer-specific
concepts (e.g., pass or dribble). One possible output of this approach are
interaction diagrams that can be used to analyse strengths of specific players
in a given match.
The approach for performance analysis in soccer in~\cite{AMS+12} also requires definition of events and situations, in this case based on a predefined ontology and rules that only use positions of objects to detect events in a game. A qualitative representation of time series in~\cite{LMV+06} has been introduced with the idea to use this representation for rule-based prediction of sequences. 

To be able to analyse, communicate, and learn team behaviour, it would be
beneficial to use an approach that also does not rely on predefined soccer
knowledge, but for example is able to learn how to describe situations or plans
from recorded logfiles of games. This way, the analysis will rely on behaviours
frequently occurring in past games instead of behaviours that may be familiar to
the developer from, e.g., human soccer matches, or similar prior knowledge that
is required when using manually specified plans.

An obvious application of identifying frequent behaviours in games is the
analysis of logfiles, to be used for programming to improve team performance, or
for automated commentator systems. A challenge for the goal of identifying
behaviours is to solve the apparently simple questions what constitutes a
behaviour and what ``situation'' means in a soccer match. Even though the
simulation naturally segments a game into steps of 100\,ms, such a fine grained
segmentation does not appear to be practical to compare longer sequences of
actions, or ``moves''.

Being able to predict likely moves from a given situation is a useful first step
towards learning behaviours for soccer players. While a few successful teams
employed machine learning techniques, most notably reinforcement learning in
e.g.,~\cite{GR06,GRT09,GR16,RM03}, or planning with MDPs in~\cite{BWC15}, use of
machine learning has been limited to components like individual skills or
behaviours.

Learning most of the teams behaviour is quite challenging due to the large
number of possible actions each player can choose from. The large number of
possible moves is also a challenge for computer programs created to play the
game of Go, and the recent success in this field~\cite{SHM+16}, for example, is
based on so-called ``policy networks''. A successful prediction of expert moves
from a given situation on the Go board has been a first step in this success
story. To take this first step in robotic soccer, the goal of our project is to
eventually be able to identify both ``actions'' and ``situations'', with the
application to learn probability distributions of the form $p(\mathrm{action} |
\mathrm{situation})$. 

\subsection{Notions}

Let us now clarify some terminology, e.g., what we mean by a situation or move.
When we analyse time-series data by machine learning techniques, we shall
distinguish the raw observed data from its reflection in a world model. In our
case, on the one hand, we have the soccer simulation data given by logfiles
representing more or less real-world data. On the other hand, we have a world
model which will be loaded into an artificial recurrent neural network,
explained later in some detail.

A soccer simulation game in the RoboCup 2D simulation league lasts
10\,mins in total, divided into 6000 cycles where the length of each cycle is
100\,ms. The central SoccerServer \cite{CDF+03} controls every virtual game with
built-in physics rules. Logfiles comprise information about the game, in
particular about the current positions of all players and the ball including
their velocity and orientation for each cycle, which we call \emph{world state}
in the following, corresponding to an actual situation.

The challenge in the simulation league is to derive for all possible world
states, possibly including their state history, the best possible action to
perform for each player, understood as intelligent agent in this context
\cite[Sect.~2]{RN09}. An \emph{action} in our case might be kick, dash (used
to accelerate the player in direction of its body), turn (change the players
body direction), and others more. Each action is executed in the actual cycle,
but its effect may hold on for a longer period of time.

For modelling the time-series induced by the logfile, we use recurrent neural
networks of a very simple form, namely echo state networks \cite{Jae07}. They consist of a number of input
neurons, containing the world state information of a given cycle. The input
neurons are connected with a reservoir usually consisting of hundreds of neurons
in a hidden layer. All neurons in the reservoir are connected randomly among
each other, at least initially. At each time point, an output signal is read out
by further connections from the reservoir to one or more output neurons. By this
procedure, each world state is reflected by a \emph{reservoir state}
corresponding to the same cycle time.

A fundamental condition for recurrent neural networks to be useful in learning
tasks is the so-called echo state property: Any random initial state of a
reservoir is forgotten, such that after a washout period the current network
state simply is a function of the driver \cite{Jae14}. We can use the series of
reservoir states for predicting future development of the game. Thus we
implicitly compute a probability distribution for the next world state or
action. In order to reach this goal, we partition the game into different moves
by clustering methods \cite[Sect. 6~and~7]{Agg15}. A \emph{move} in this
context means a pattern of similar states. It is derived from a sequence of
consecutive reservoir states which might occur several times during the whole
game but it means the respective series of world states.

The idea is that each move corresponds to a certain behaviour such as e.g.
kicking a goal. When a recurrent neural network is actively generating or is
passively being driven by different dynamical patterns, its reservoir states are
located in different regions of the multi-dimensional neural state space,
characteristic for that pattern. A move thus corresponds to a specific concept
which can be captured by a \emph{conceptor} \cite{Jae14}. They allows us to
analyse and identify moves and finally to predict the development of the game.
This is the main objective of this paper.

\subsection{Overview}

The rest of the paper is structured as follows: Next, we will describe the used
methods in more detail, from data preparation to recurrent neural networks with
conceptors and clustering algorithms (Sect.~\ref{sec:methods}). After that, we
state some results on clustering and prediction in soccer games
(Sect.~\ref{sec:results}). Finally, we and up with conclusions and a brief
outlook on future work (Sect.~\ref{sec:final}).

\section{Methods}\label{sec:methods}

\subsection{Data Preparation}

In order to prepare the data needed, we use recorded logfiles from 2D soccer
simulation matches between two teams. The main data set we will be using
throughout this paper is a match between the teams Gliders2012~\cite{POWH12} and
MarliK~\cite{TNV+12}.\footnote{The logfile is taken from the site
\href{http://chaosscripting.net/}{http://chaosscripting.net/}, a backup
of the official soccer simulation website.}
For more detailed experiments, we also created a data set
of 101 games, run using the teams Helios2016~\cite{ANH+16} and
Gliders2016~\cite{PWOJ16}. Each logfile contains the ``state of the world'',
ground truth information sampled every 100\,ms, which includes the positions of
the players and the ball in $x$ and $y$ coordinates, as well as the speed and
the stamina of the players.
 
The soccer simulator records games in a binary format, which we converted to
text using the tool \textsf{rcg2txt}, from the \textsf{librcsc}
library~\cite{AS11}. The resulting data were then converted from text to csv
files (comma separated values) using a custom Python script. For the work in
this paper, we only made use of player and ball positions (23 objects with $x$-
and $y$-positions, i.e. 46 values for each step of the simulation). As already
said, each match lasts 6000 steps (10\,mins in real time), but in some
situations game time may be stopped, which can lead to extra steps. On average,
we recorded close to 6500 game states per match. 

\subsection{Recurrent Neural Networks with Conceptors}

\begin{figure}
\centering
  \includegraphics[width=0.7\textwidth]{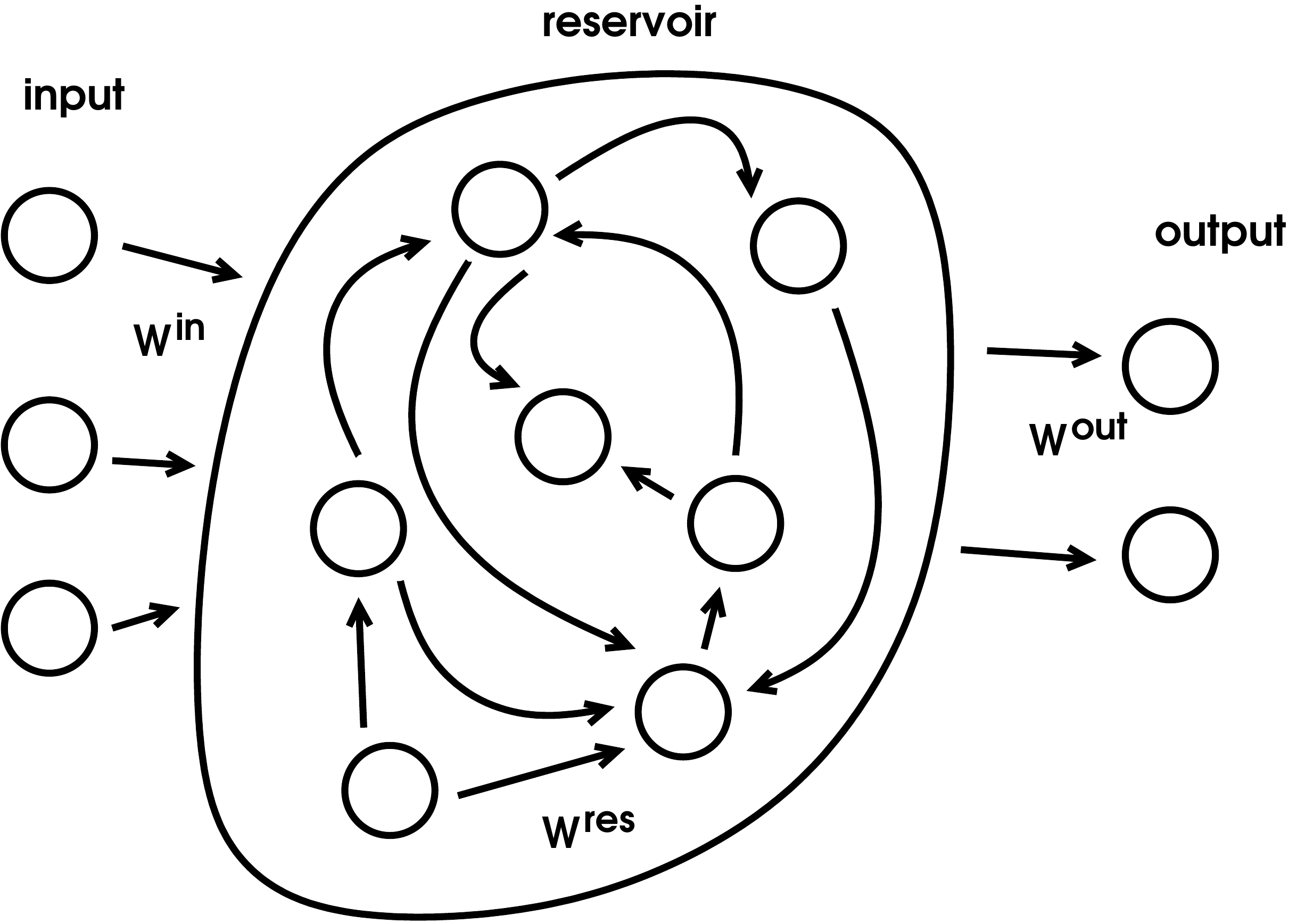}
\caption{Recurrent neural network capturing the world model.}\label{fig:network}
\end{figure}

The world model of the time-series data, as created in the data preperation phase, is loaded into a recurrent neural
network. Following the lines of \cite[Sect.~3]{Jae14}, a (discrete-time)
\emph{recurrent neural network} consists of (a) $N^\mathrm{in}$ input neurons,
(b) a reservoir, that is a set of $N$ recurrently connected neurons, and (c)
$N^\mathrm{out}$ output neurons (cf.\ Fig.~\ref{fig:network}). The connections between
neurons have weights which may be comprised in weight matrices: $W^\mathrm{in}$
is the $N \times N^\mathrm{in}$ input weight matrix, $W^\mathrm{res}$ the $N \times N$
reservoir matrix, and $W^\mathrm{out}$ the $N^\mathrm{out} \times N$ readout
matrix. The network operates in discrete time steps $n=0,1,2,\dots$ according to
the following update equations:
\begin{eqnarray*}
x(n+1) & = & \tanh(W^\mathrm{res} \cdot x(n) + W^\mathrm{in} \cdot p(n+1) +b)\\
y(n) & = & W^\mathrm{out} \cdot x(n)
\end{eqnarray*}

Here $p(n)$ denotes the input (training) signal (here: the logged data from the
recorded soccer games), $x(n)$ the reservoir state, and $y(n)$ the target signal
(here: the next input state). $b$ is an $N \times 1$ bias vector. Like the input
weights, it is fixed to random values. Due to the commonly used non-linear
activation function $\tanh$, the reservoir state space is reduced to the
interval $(-1;+1)$. A time-series may represent a move, which can be learned by
so-called conceptors as follows: The sequence of world states representing the
move is stored into the reservoir such that it can generate the driven responses
$x(n)$ even in the absence of the driving input. For this, ridge regression may
be employed. Primarily, the output weights $W^\mathrm{out}$ are learned.
Nevertheless, in order to retrieve the output pattern individually,
the reservoir dynamics is restricted to the linear subspace characteristic for
that pattern. This is done by special matrices called conceptors. These are
lower-dimensional manifolds that describe trajectories through a
high-dimensional state space that enable situation-specific predictions from the
same neural network. For more
details the reader is referred to \cite{Jae14}.

For the analysis of soccer games, different moves are associated with
conceptors. However, we usually do not know in advance which are the moves of
the whole game. Thus we first have to establish a repertoire of moves that form
the whole game. We want to identify and learn moves by clustering methods, which
is explained in the next section. To support a wide set of strategies, our goal
is to identify situations from data, in an unsupervised way without making use
of pre-defined soccer specific concepts such as ``pass'' or ``dribble''.

\subsection{Clustering of Games in Moves}

\begin{figure}
\includegraphics[width=\textwidth]{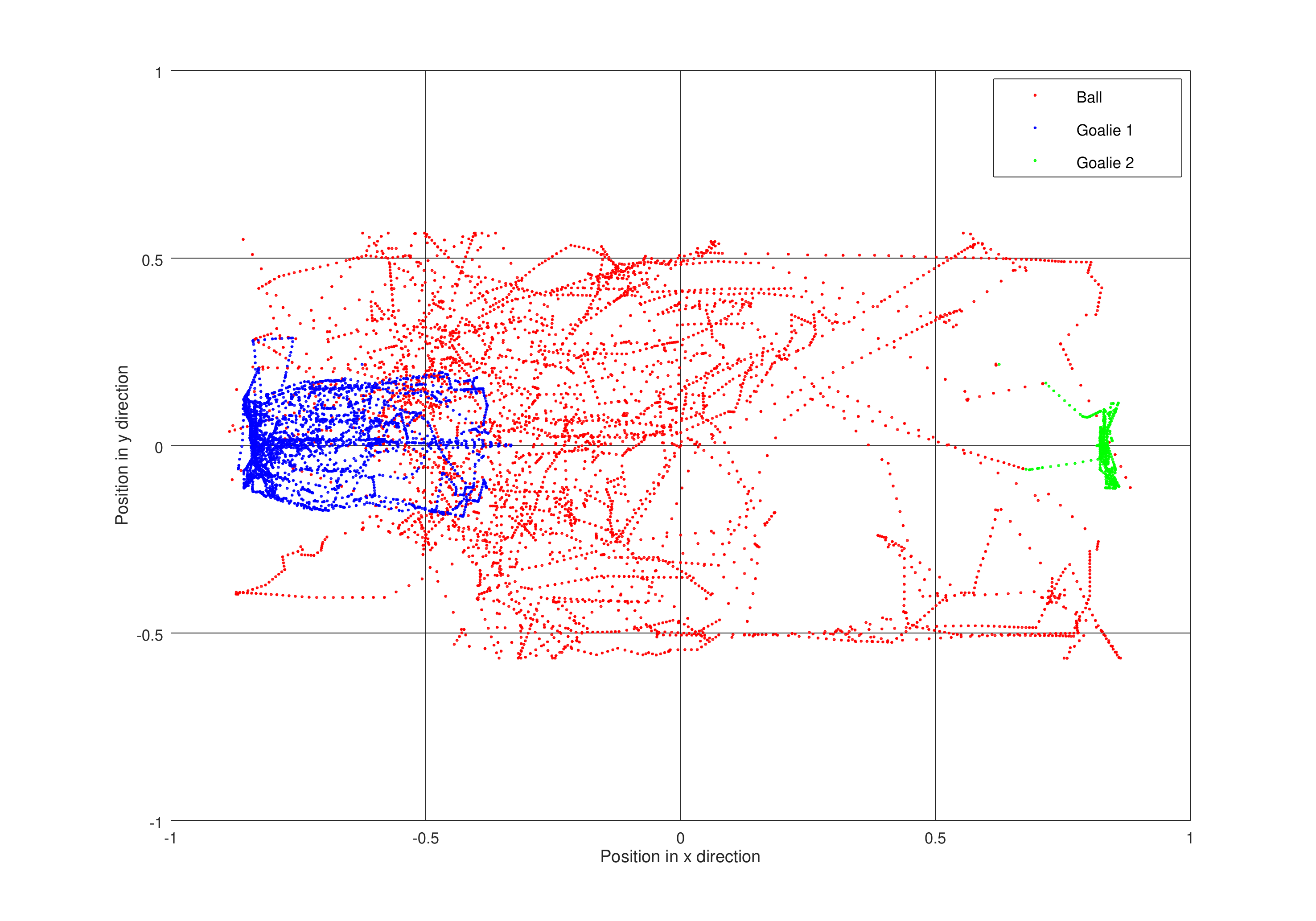}
\caption{Position visualization of the two goalies and the ball.}\label{fig:gamelogfileviz}
\end{figure}

As described above, we extract the positions of all players and the ball from logfiles, 
for each cycle of the game and store them as world states normalised to the range $(-1;+1)$ (because of the use of the $\tanh$ function, see above).
In the example in Fig.~\ref{fig:gamelogfileviz}, all world states from a logfile
MarliK vs. Gliders2012 are drawn, here for the ball and the two goalies.

To identify the sequences of world states that represent moves leading to
similar game situations, we use the actual world state $p(i)$ to compute the reservoir
state $x(i)$ for each cycle of the game. 
A reservoir state represents the current state of the game, as well as the 
history that led to that state, therefore we can assume that similar reservoir states 
represent  similar moves in a game. Clustering of all reservoir states $x(i)$ is able to 
identify such similar moves. Cluster members are close to each other in the reservoir state 
space; that is, for each member the current game situation and its short-term past are 
similar to other cluster members. 

We use the X-means algorithm \cite{PellegM00} to derive the clusters and the number of clusters. This algorithm extends
the well-known $k$-means clustering method with an efficient estimation of the
number of clusters. While the number of clusters has to be fixed in advance in
the $k$-means methods, X-means clustering starts with $k=2$ clusters and splits
clusters, if their average squared distance to the centroid still can be
improved.

\section{Results}\label{sec:results}

To test our approach we extract the 6493 world states from the logfile
MarliK\_1-vs-Gliders2012\_3 and store these states into a recurrent neural
network with 46 input neurons, 600 neurons in the reservoir and 46 output neurons.
We compute the reservoir states $x(i)$ for each world state and cluster them with the
X-Means algorithm of the data mining tool WEKA~3.8 \cite{Frank2016,hall09}.
With the maximal number of clusters of 100 and a maximum of 5 iterations as presetting parameters,
we get 64 clusters.

By this procedure, we achieved two main results: First, each resulting cluster
can be associated with one or more sequences of world states, i.e. with moves.
Each cluster consists of a set of similar reservoir states. The reservoir states
in each cluster may come from different phases of the soccer game. Nevertheless in a cluster
often several reservoir states stem from consecutive time steps. The reason for
this is that the world state and hence the reservoir state does not change too much from one time step to
the other according to the network dynamics. Each sequence of these consecutive
reservoir states now belongs to a move. Note that in the figures only the final phases of
the moves, strictly speaking the world states corresponding to the consecutive
reservoirs states, are shown. Actually the moves start some time steps earlier.

\begin{figure}
\includegraphics[width=\textwidth]{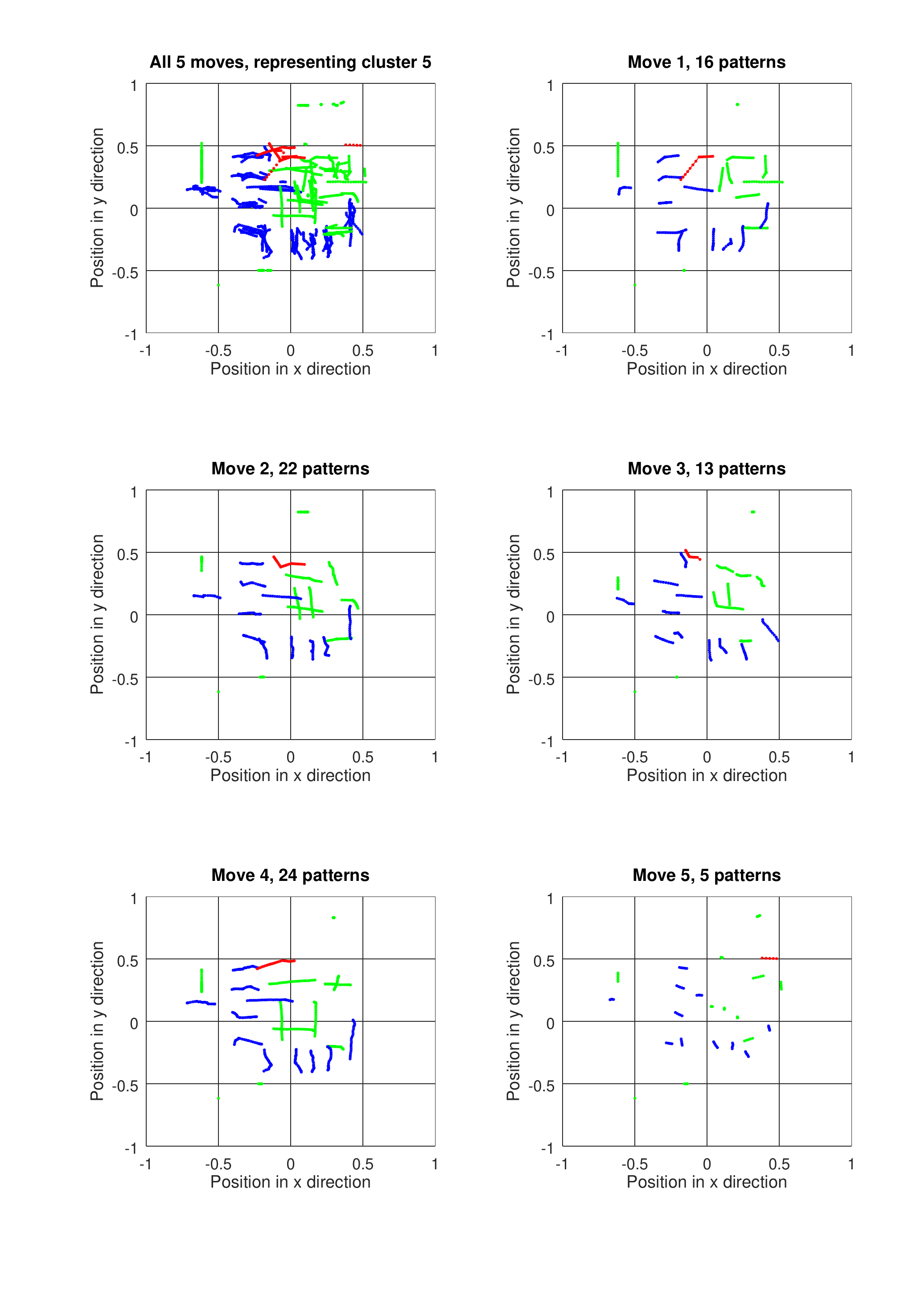}
\caption{All states referring to cluster 5. The trajectories of the ball (red), of the left team (blue), and of the right team (green) are shown (colours in online version).}\label{fig:clusterexample}
\end{figure}

\begin{figure}
\includegraphics[width=\textwidth]{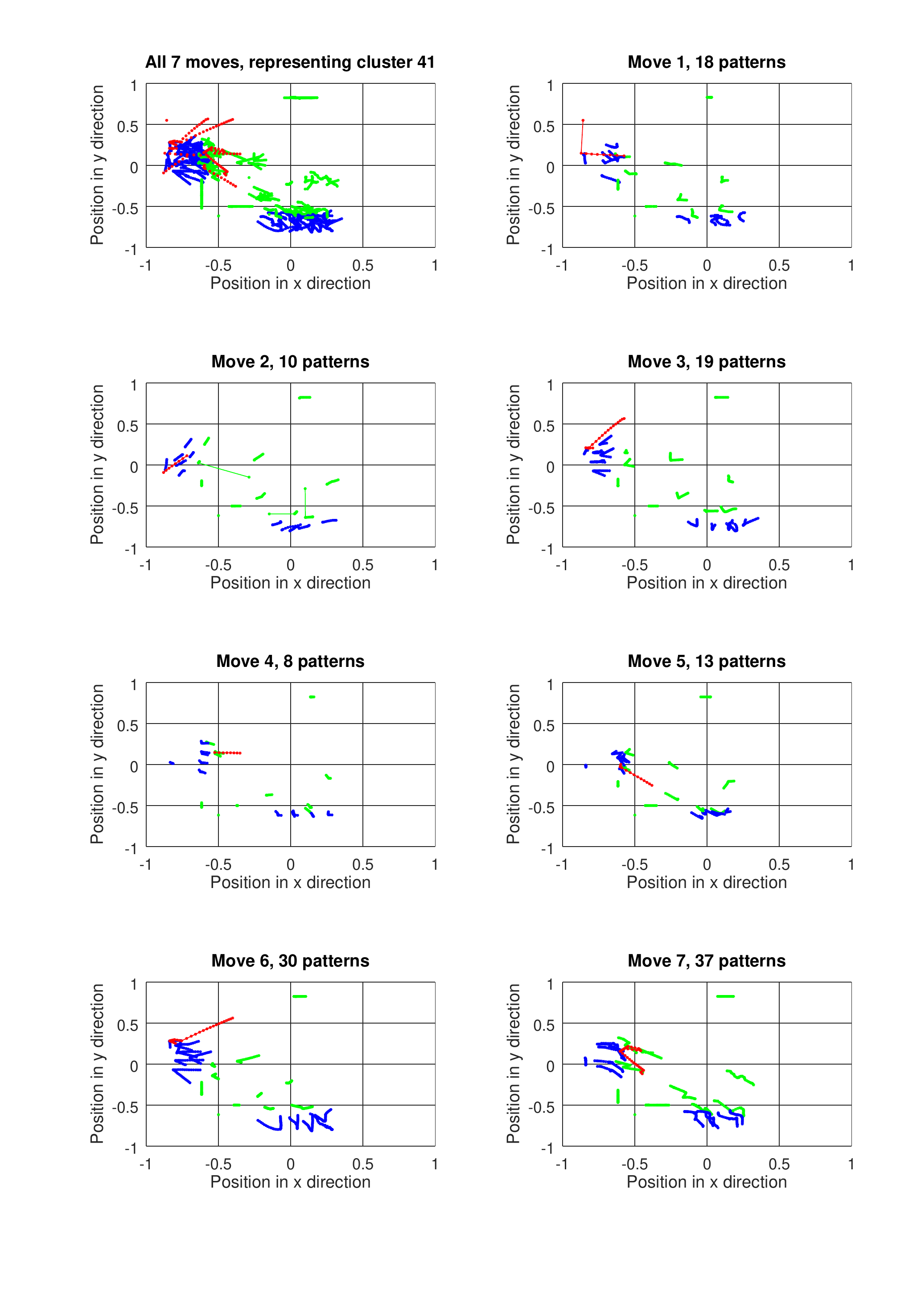}
\caption{All states referring to cluster 41. Again, the trajectories of the ball (red), of the left team (blue), and the right team (green) are shown (colours in online version).}\label{fig:clusterexample2}
\end{figure}

Figure~\ref{fig:clusterexample} shows all 5 moves of cluster 5 for all players and
the ball, and Fig.~\ref{fig:clusterexample2} all moves of cluster 41. In the
subfigures for each move, the part of the respective world state sequence is
drawn, where the reservoir states $x(i)$ are in the actual cluster. With the now
identified clusters and moves it is possible to compute the conceptors for these
moves in the respective game situations. These conceptors allow us to predict
the next world states in similar situations in the game, e.g., whether the next
action will be a kick to the goal.

\begin{figure}[t]
\centering
\includegraphics[width=0.9\textwidth]{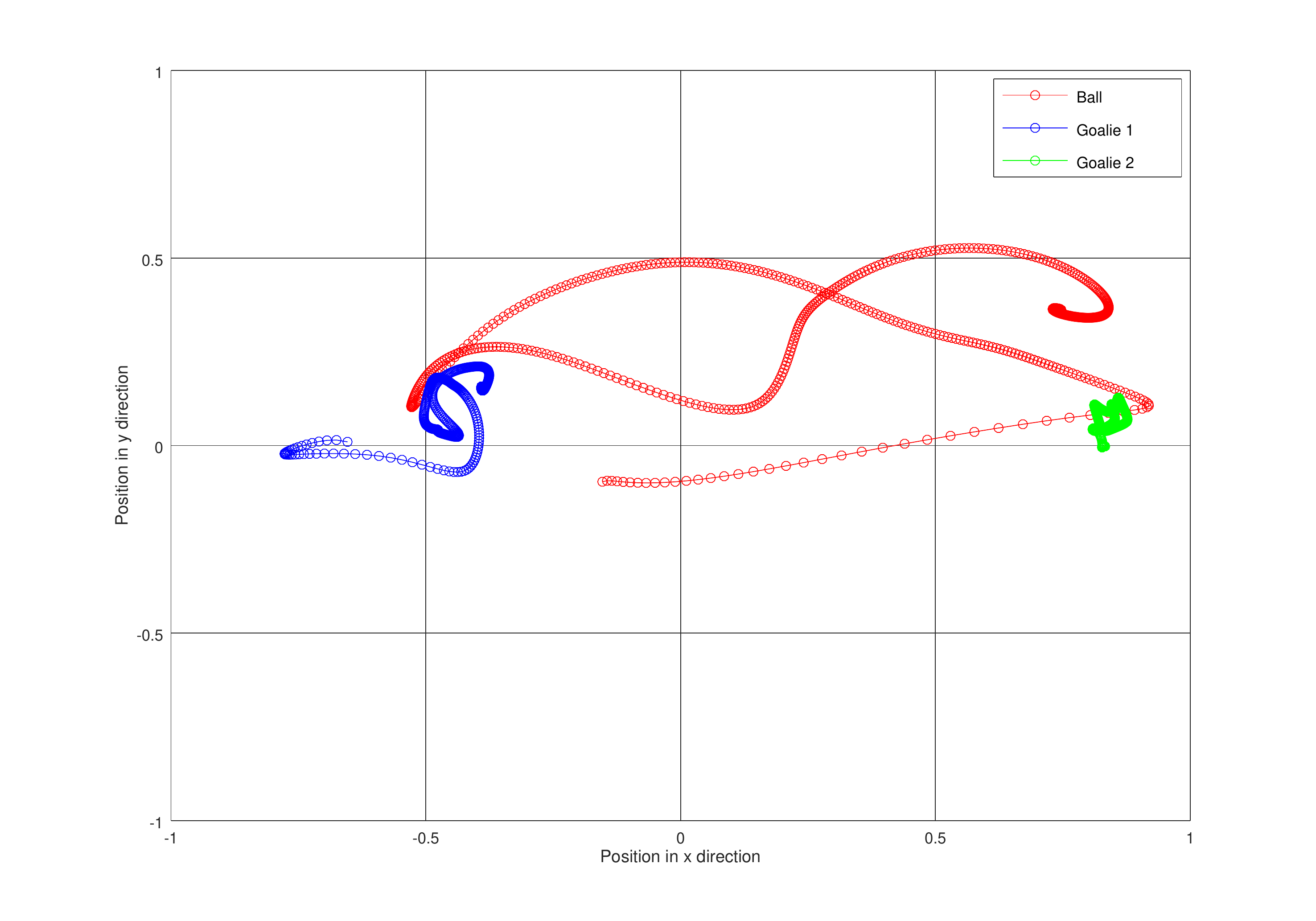}
\caption{Prediction of an entire game by the reservoir dynamics of the recurrent neural network.}\label{fig:gameprediction}
\end{figure}

As a second result, with the resulting conceptor of the reservoir filled with
all world states, i.e. over all time steps of the whole game, we can also
respresent a simplified picture of the entire game, i.e. a kind of replay. This
is shown in Fig.~\ref{fig:gameprediction}. For the sake of simplicity, only
goalies and ball are drawn here. As one can see, the figure shows that the
goalies keep to the goal area and that the ball moves essentially back and forth
from one goal to the other. This seems to be an absolutely plausible summary of
the whole soccer game. Of course, this point needs further investigation.

\section{Discussion and Final Remarks}\label{sec:final}

We have shown how clustering methods in conjunction with conceptors in recurrent neural networks help to
determine moves of soccer game. Conceptors allow us to make predictions of the
whole game or the next move, e.g., whether there is a probability of kicking a
goal.

Future work will concentrate on analysing more deeply different moves and their
structure and the prediction accuracy of conceptors. The overall method shall be
improved such that it is applicable in real-time and online. For this, the
neural network model shall be simplified even further.

\subsubsection*{Acknowledgements.}
The research reported in this paper has been supported by the German Academic
Exchange Service (DAAD) by funds of the German Federal Ministry of Education and
Research (BMBF) in the Programmes for Project-Related Personal Exchange (PPP)
under grant no.~57319564 and Universities Australia (UA) in the
Australia-Germany Joint Research Cooperation Scheme within the project \emph{\underline{De}ep
\underline{Co}nceptors for Tempo\underline{r}al D\underline{at}a
M\underline{in}in\underline{g}} (Decorating).

\end{document}